%% file: main.tex
\RequirePackage{fix-cm}
\documentclass[twocolumn]{svjour3}          
\smartqed  
\usepackage{graphicx}
\usepackage{natbib}
\usepackage{multirow}
\usepackage{amsmath}
\usepackage{amssymb}
\usepackage{bbm}
\usepackage{bm}
\usepackage{booktabs}
\usepackage{array} 
\usepackage{blindtext}
\usepackage{comment}

\usepackage{tikz}
\usepackage{textcomp}
\usepackage{lipsum}
\usepackage{algorithm}
\usepackage[noend]{algpseudocode}
\usepackage{bigstrut}
\usepackage{listings} 
\usepackage{pythonhighlight}
\newcommand{\para}[1]{\textbf{#1}}

\usepackage{fontawesome}

\usepackage{pifont}

\usepackage{color, colortbl}
\definecolor{citecolor}{HTML}{0071bc}
\definecolor{tabhighlight}{HTML}{e5e5e5}
\usepackage[colorlinks=true,linkcolor=blue,citecolor=citecolor,urlcolor=blue]{hyperref}

\makeatletter
\renewcommand\paragraph{
  \@startsection{paragraph} 
  {4} 
  {\z@} 
  {.5em \@plus1ex \@minus.2ex} 
  {-.5em} 
  {\normalfont\normalsize\bfseries} 
}
\makeatother
\begin{document}
\sloppy

\title{Progressive Volume Distillation with Active Learning for Efficient NeRF Architecture Conversion}
\newcommand{\modelname}{CE3D}

\author{
	Shuangkang Fang $^1$\and
	Yufeng Wang$^2$\textsuperscript{\faEnvelopeO}\and 
	Yi Yang$^3$\and
        Weixin Xu$^3$\and \\
        Heng Wang$^3$\and
        Wenrui Ding$^2$\and
        Shuchang Zhou$^3$
}

\institute{
$^1$ School of Electrical Information Engineering, Beihang University, Beijing, China \\
$^2$ Institute of Unmanned System, Beihang University, Beijing, China \\
$^3$ Megvii Research, Megvii Inc., Beijing, China \\
\textsuperscript{\faEnvelopeO} Corresponding author: wyfeng@buaa.edu.cn
}


\maketitle

\begin{abstract}
Neural Radiance Fields (NeRF) have been widely adopted as practical and versatile representations for 3D scenes, facilitating various downstream tasks. However, different architectures, including the plain Multi-Layer Perceptron (MLP), Tensors, low-rank Tensors, Hashtables, and their combinations, entail distinct trade-offs. For instance, representations based on Hashtables enable faster rendering but lack clear geometric meaning, thereby posing challenges for spatial-relation-aware editing. To address this limitation and maximize the potential of each architecture, we propose Progressive Volume Distillation with Active Learning (PVD-AL), a systematic distillation method that enables any-to-any conversion between diverse architectures. PVD-AL decomposes each structure into two parts and progressively performs distillation from shallower to deeper volume representation, leveraging effective information retrieved from the rendering process. Additionally, a three-level active learning technique provides continuous feedback from teacher to student during the distillation process, achieving high-performance outcomes. Experimental evidence showcases the effectiveness of our method across multiple benchmark datasets. For instance, PVD-AL can distill an MLP-based model from a Hashtables-based model at a 10$\times$$\sim$20$\times$ faster speed and 0.8dB$\sim$2dB higher PSNR than training the MLP-based model from scratch. Moreover, PVD-AL permits the fusion of diverse features among distinct structures, enabling models with multiple editing properties and providing a more efficient model to meet real-time requirements like mobile devices. 
\end{abstract}

\section{Introduction} \label{sec:introduction}

\begin{figure*}[t]
  \vspace{-2 ex}
        \centering
        \includegraphics[width=0.98\textwidth]{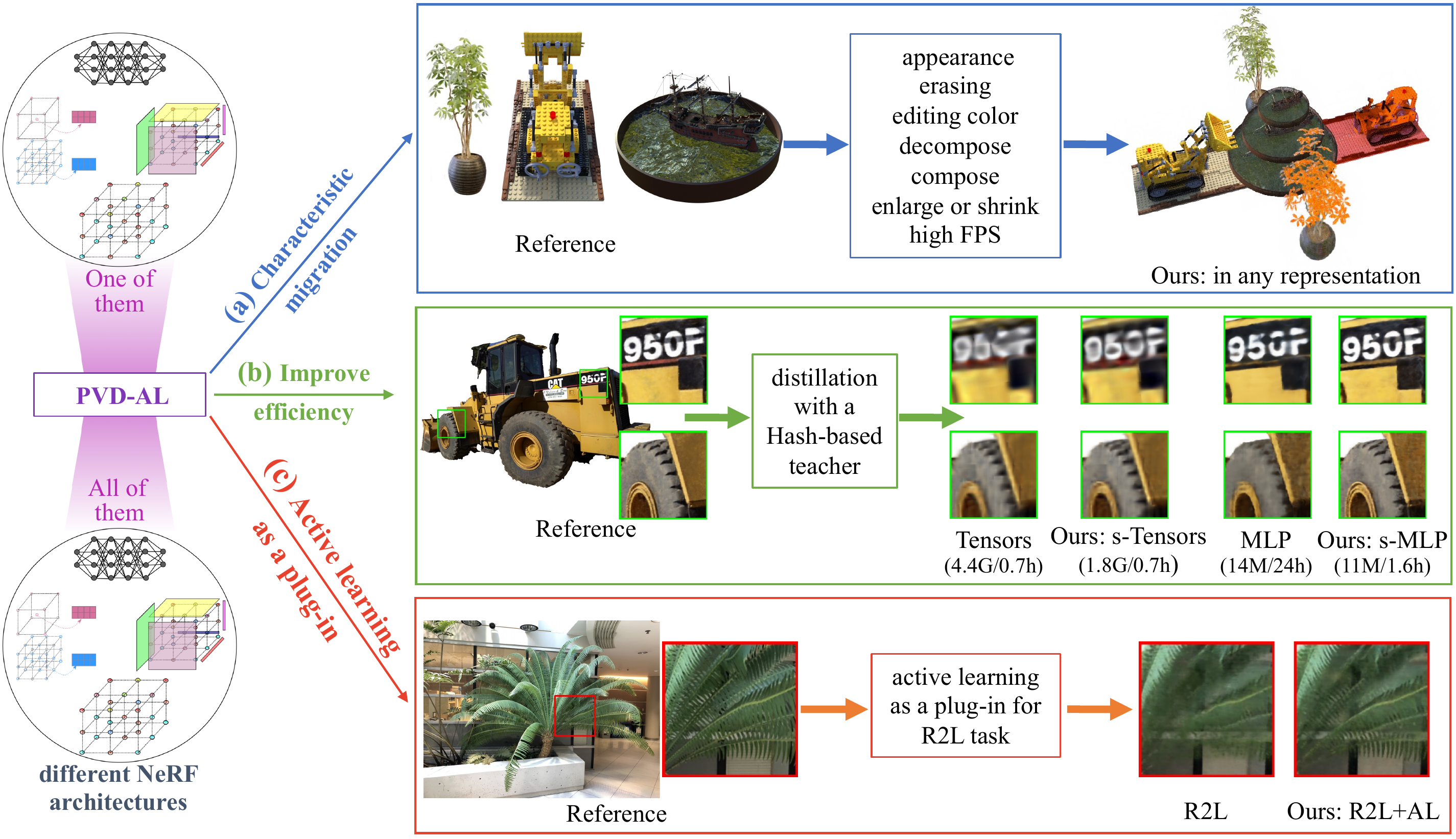}
        \caption{\textbf{Capabilities of the proposed PVD-AL method}. PVD-AL facilitates different NeRF architecture conversions, breaking down the barrier of independent research between them. Its efficient and flexible design supports versatile applications. (a) PVD-AL has the ability to migrate any architecture's attributes to other architectures or concentrate them on a single architecture. (b) A model obtained by PVD-AL performs better than by training it from scratch, with less parameters and training time. (c) Active learning strategy can be plugged into other NeRF-related distillation tasks (like radiance field to light field: R2L\citep{wang2022r2l}) to enhance final performance. More results can be found in Section~\ref{sec:experiments}.}
        \label{fig:shocking}
  \end{figure*}

Novel view synthesis (NVS) generates photo-realistic 2D images for unknown view-ports of a 3D scene~\citep{zhou2018stereo,chan2021pi,sitzmann2019scene}, which has wide applications in rendering, localization, and robot arm manipulations~\citep{adamkiewicz2022vision,moreau2022lens,peng2021megloc}, especially with the neural modeling capabilities offered by the NeRF~\citep{mildenhall2020nerf}.
By exploiting the implicit modeling capabilities of MLP, NeRF can significantly improve the quality of NVS.
Several following developments incorporate feature tensors as complementary explicit representations to relieve the MLP from remembering all details of the scene, resulting in faster training speed and more flexible manipulation of geometric structure. The bloated size of the feature tensors in turn spurs works targeting more compact representations, like Plenoxels~\citep{fridovich2022plenoxels} that exploits the sparsity of the tensor, TensoRF~\citep{chen2022tensorf} that leverages VM (vector-matrix) decomposition and canonical polyadic decomposition (CPD), and Instant Neural Graphics Primitives (INGP)~\citep{muller2022instant} that utilizes multilevel Hashtables for effective compression of feature tensors.
All these schemes have their own advantages and limitations. 
Generally, with implicit representations, it would be easier to perform texture editing, artistic stylization, and dynamic scene modeling~\citep{fang2023dn2n,kobayashi2022decomposingEditing,wang2024uavenerf,gu2021styleNeRF,li2022climatenerf}. 
On the other hand, methods with explicit representation usually enjoy faster training due to the shallower representations and cope better with geometric-aware editing~\citep{tang2022CCNeRF, yu2021plenoctrees, fridovich2022plenoxels}, like merging and other manipulations of scenes, which is in clear contrast to the case of purely implicit representations. Besides, the compatibility of different architectures with various hardware platforms can vary substantially.

Due to the diversity of downstream tasks of NVS, there is \textit{no single answer} as to which representation is the best. The particular choice would depend on the specific application scenarios and the available hardware resources. \textit{Researching them independently can not reach their full potential}.
In this paper, we tackle the problem from another perspective. Instead of focusing on an ideal representation that embraces the advantages of all variants, we propose a method to achieve arbitrary conversions between known NeRF architectures, including MLP, sparse Tensors, low-rank Tensors, Hashtables, and combinations thereof. Such flexible conversions can obviously bring the following advantages:
(1) The study would throw insights into the modeling capabilities and limitations of the already rich and ever-growing constellation of architectures of NeRF.
(2) The possibility of such conversions would free the designer from the burden of pinning down architectures beforehand, as now they can simply adapt a trained model agilely to other architectures to meet the needs of later discovered application scenarios. 
(3) Attributes between different structures can complement each other. For instance, the implicit MLP can inherit spatial-aware editing capabilities from the explicit Tensors.
(4) This research offers the opportunity to transfer knowledge of high-quality 3D representations to downstream tasks, demonstrating significant practical value.

In order to convert between various architectures, two issues must be addressed. One is that there are significant differences in training and inference speeds across different architectures. How can we ensure that the slower model does not hinder conversion efficiency? The second is how to condense the most relevant and useful information from a well-trained model and transfer it to another model.

We propose PVD-AL as a solution to address the two issues noted above. It decomposes each architecture into two parts and applies the effective information retrieved from the rendering process to perform the distillation process progressively, stage-by-stage, on different levels of volume representation, from shallower to deeper, thus accomplishing the objective of rapid distillation. 
In this design, the teacher directs student training with the high-level semantic information of the model's middle layer, the density and color of the spatial points, and the rendered RGB values. 

However, we find that relying solely on the techniques mentioned above may not be adequate for effective learning.
For example, some sampling rays that pass through blank areas can be easily fitted by students, whereas for rays that intersect object surfaces, students typically require more time to fit. Similar rules apply to camera poses and sample points along a ray. 
Therefore, we introduce an active learning strategy in PVD-AL. By including this strategy in the distillation procedure, students would be informed in real-time which camera poses, sample rays and sample points offer the greatest challenge. Then students will pay more attention to these vital pieces of knowledge after obtaining this feedback in the next training step, leading to high-performance distillation results.

To sum up, the contributions of this paper are summarized as follows:

\begin{itemize}
\item We propose PVD-AL, a distillation framework to accelerate the training procedure based on a unified view that allows conversions between different NeRF architectures, including MLP, sparse Tensors, low-rank Tensors, and Hashtables. To the best of our knowledge, this is the first systematic attempt at such conversions.

\item We introduce an active learning strategy so that students can acquire knowledge from teachers to the greatest extent. We continuously evaluate the camera poses, sample rays, and sample points that are difficult to fit for students, which help them actively enhance the learning of crucial knowledge. The three-level active learning strategy is decoupled and flexible; thus can also be easily applied as a plug-in to other distillation tasks based on NeRF.

\item Using high-performance teachers, such as Hashtables and VM-decomposition structures, frequently improves student model synthesis quality while taking less time than training the student from scratch. Consequently, our technique serves as a valuable tool for model training and compression, yielding more effective models for various applications.

\item PVD-AL allows for the fusion of various properties between different structures. For example, we can call PVD-AL multiple times to obtain models with diverse editing properties or to convert a scene under a specific model to another one that runs more efficiently to meet real-time requirements and enrich scene content on mobile devices.
\end{itemize}

This article constitutes a substantial extension of our previous work~\citep{fang2022pvd}, differing primarily in four aspects: (1) We introduce a novel three-level active learning strategy to fully exploit the potential for transformation between different NeRF structures and conduct entirely new mutual-conversion and visualization experiments to validate the effectiveness of the new method. 
(2) We thoroughly explore the capabilities of PVD-AL, including the transfer of editing abilities across different architectures, model compression, model acceleration, and as a plugin to enhance the performance of other NeRF-based distillation methods. (3) We corroborate PVD-AL's value in practical applications by obtaining 3D models operable in real-time on mobile devices and expanding the diversity of scenes on such devices. (4) We conduct ablation experiments to investigate the importance of the various components introduced in the new method, and further present and analyze the results of fine-tuning after distillation.

\begin{figure*}[t]
  \vspace{-2 ex}
        \centering
        \includegraphics[width=1.\textwidth]{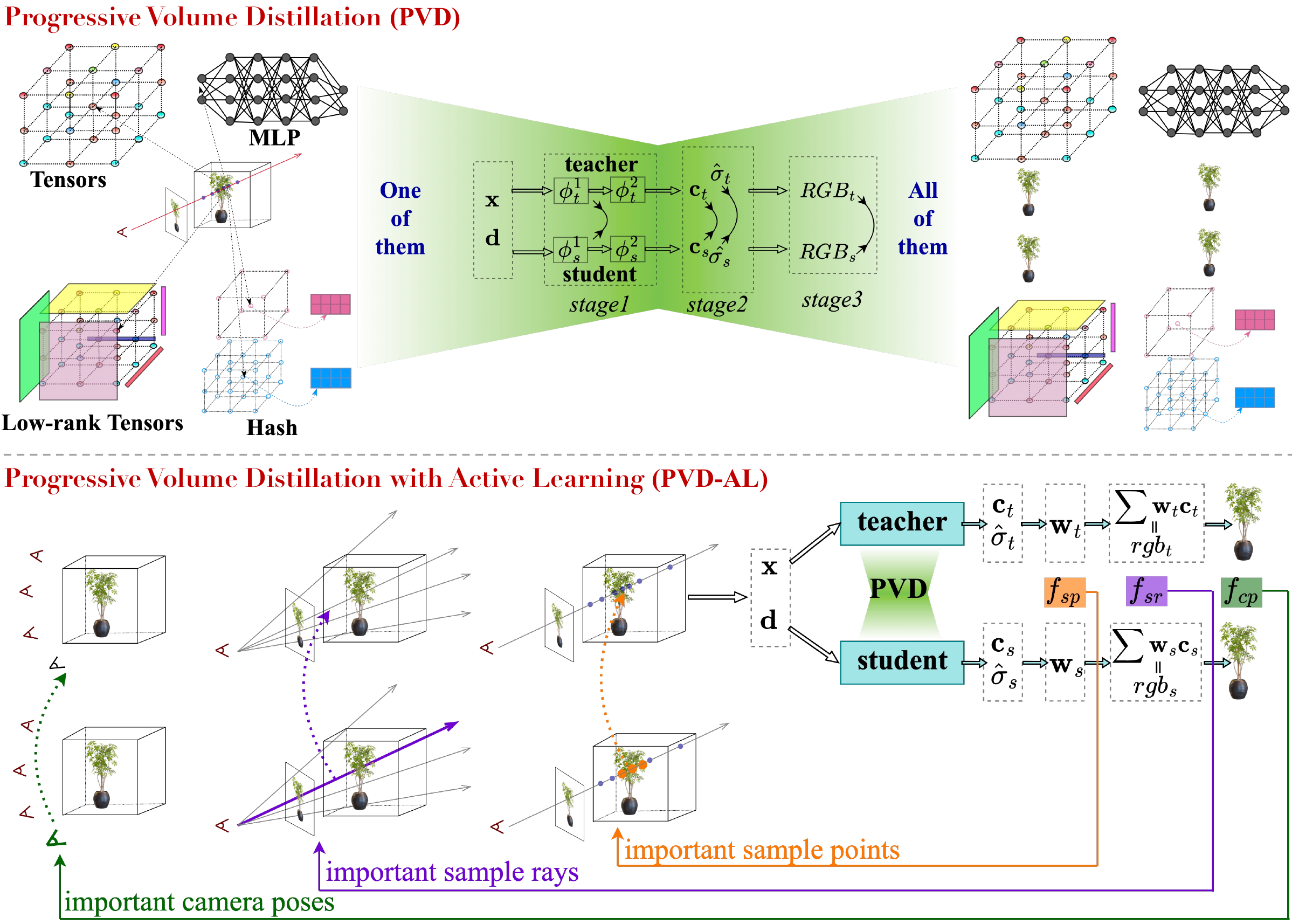}
        \caption{\textbf{Illustration of the PVD-AL framework}. Given one trained NeRF model, different NeRF architectures, such as the MLP, sparse Tensors, low-rank Tensors, and Hashtables can be quickly obtained through PVD-AL. The loss in intermediate volume representations (shown as a double arrow symbol in PVD) like the output of $\phi_{*}^1$, color, and density are used alongside the final rendered RGB volume to accelerate distillation. In PVD-AL, regular feedback from the teacher will be given to the student. This form of feedback is organized into three tiers based on camera poses, sample rays, and sample points, which prepares the student to proactively learn what they need to be strengthened for the next round of training.}
        \label{fig:pvd}
  \end{figure*}

\section{Related Work}
\subsection{Neural Implicit Representations} 

Neural implicit representation methods use MLP to construct a 3D scene from coordinate space, as proposed in NeRF~\citep{mildenhall2020nerf}. The input of the MLP is a 5D coordinate (spatial location [$x, y, z$] and viewing direction [$\theta, \phi$]), and the output is the volume density and color.
An advantage of implicit modeling is that it is conducive to controlling or changing the texture-like attributes of the scene~\citep{liu2023stylerf,wang2022clipnerf}. For example, DFFs~\citep{kobayashi2022decomposingEditing} use the pretrained CLIP model~\citep{radford2021clip} to induce editing of NeRF representation of a scene. 
DoReNeRF~\citep{lou2024darenerf} successfully applies NeRF to the rendering of dynamic scenes by mapping time $t$ to implicit space through an MLP. NeRFW~\citep{martin2021nerfinthewild} realizes the control of scene lighting by adding appearance embedding using an MLP. However, the MLP-based model requires on-the-fly dense sampling of spatial points, which leads to multiple queries of the MLP during training and inference, resulting in slower running speed.

\subsection{Neural Explicit Representations and Hybrids}
With explicit representations, a scene is placed directly on a 3D grid. Each voxel on the grid stores information on density and color. Plenoxels~\citep{fridovich2022plenoxels} shows that a 3D scene can be represented by an explicit grid, with the spherical harmonic coefficients at each voxel enabling the retrieval of density and color values at any spatial point via trilinear interpolation.
The training and inference speed of Plenoxels is significantly superior to that of MLP-based NeRF. 
Motivated by the low-rank tensor approximation algorithm, TensoRF~\citep{chen2022tensorf} decomposes the explicit tensor into low-rank components, which significantly reduces the model size. 
PeRF~\citep{rasmuson2022perf} continues to evolve the explicit expression and regard the optimization of the grid as a non-linear least squares optimization problem that can be solved more efficiently by the Gauss-Newton method. 
With explicit representation, it is not as easy to make artistic creations as with implicit representation. Nevertheless, explicit representations facilitate the geometry editing of the scene, including merging of multiple scenes, inpainting, and manipulations of objects at specific positions~\citep{tang2022CCNeRF, yu2021plenoctrees, fridovich2022plenoxels}.

There are also attempts exploiting a hybrid of the explicit and implicit representations as NeRF architectures~\citep{wang2024uavenerf,zhang2023efficientgpnerf,usvyatsov2022t4dt,muller2022instant,chen2022tensorf,wu2022diver}.
The explicit part usually stores features related to the scene, while the implicit part is typically an MLP that interprets the features to obtain densities and colors. Differences between hybrid representations are mainly exhibited in the explicit part. NSVF~\citep{liu2020nsvf} uses a spare grid to store features, while Plenoctrees~\citep{yu2021plenoctrees} optimizes the 3D grid through an octree. Nex~\citep{wizadwongsa2021nex} proposes an Implicit-Explicit modeling strategy by storing the coefficient as a learnable parameter to accelerate the training procedure. Recently, INGP~\citep{muller2022instant} proposes the multi-resolution hash encoding (MHE), which maps the given coordinate to a feature via a cascade of Hashtables at different scales. Like TensoRF~\citep{chen2022tensorf}, MHE significantly reduces memory footprint and improves inference speed. However, the compactness of MHE comes at the cost of less straightforward geometric interpretation, as there are abundant spatial aliases caused by the hash mechanism.

\subsection{Knowledge Distillation}
Knowledge distillation commonly refers to training a small model to match the output of a larger model (maybe trained beforehand or on the fly), which is widely used in model optimization and compression~\citep{hinton2015distilling,huang2024knowledge,hao2024one-distillation,leng2023dffg}. Multiple attempts have been made in the field of NVS. Mip-nerf 360~\citep{barron2022mip} proposes an online distillation method to improve the quality of rendering. R2L~\citep{wang2022r2l} and MobileR2L~\citep{cao2022mobiler2l} convert a NeRF model into a model based on neural light fields. The most related to our work is KiloNeRF~\citep{reiser2021kiloNeRF}, which uses a huge pretrained NeRF (teacher) to guide thousands of small NeRF models (students) for speeding up. However, KiloNeRF only performs distillation between the same MLP architecture and the distilling process is significantly slowed down by the continuous querying of the huge MLP in the teacher model.

\subsection{Active Learning}
Active learning is a special case of machine learning in which a learning algorithm can interactively query a teacher to label new data points with the desired outputs\citep{active2021survey,tharwat2023survey-active,manjah2023-active1,boreshban2023-active2}. There are situations in which unlabeled data is abundant but manual labeling is expensive. 
In such a scenario, learning algorithms can actively query the user or teacher for labels. This type of iterative supervised learning is called active learning. In such a learning process, the learner selects examples, potentially reducing the number needed to grasp a concept compared to traditional supervised learning~\citep{tharwat2023-active3,rangnekar2023active4,gao2020-active5}. 
ActiveNeRF\citep{pan2022activenerf} proposes to supplement the existing training set with newly captured samples by camera based on an active learning scheme, which is totally different from our PVD-AL method.

\section{Method}
Our method aims to achieve mutual conversions between different NeRF architectures. Since there is an ever-increasing number of such architectures, we are not attempting to achieve these conversions one by one. Rather, we first formulate typical architectures in a unified form and then design a systematic distillation scheme based on the unified view. These architectures include implicit representations like MLP in NeRF, explicit representations like sparse Tensors in Plenoxels, and two hybrid representations: Hashtables (in INGP) and low-rank Tensors (VM-decomposition in TensoRF).
We first cover some preliminary topics in Subsection~\ref{sec:preliminaries},
then moving on to a detailed description of PVD-AL in Subsection~\ref{sec:pvd-al}, including the loss design, density range constraint, block-wise distillation, and three levels of active learning.

\subsection{Preliminaries} \label{sec:preliminaries}
\para{Neural Radiance Fields.}
NeRF represents scenes with an implicit function that maps spatial point $\mathbf{x}=(x, y, z)$ and view direction $\mathbf{d}=(\theta, \phi)$ into the density $\sigma$ and color $\mathbf{c}$. Given a ray $\mathbf{r}$ originating at $\mathbf{o}$ with direction $\mathbf{d}$, the RGB value $\hat{\mathbf{C}}(\mathbf{r})$ of the corresponding pixel is estimated by the numerical quadrature of the color $\mathbf{c}_i$ and density $\sigma_i$ of the spatial points $\mathbf{x}_i=\mathbf{o}+t_i\mathbf{d}$ sampled along the ray:
\begin{equation}  \label{neural_rendering_equation}
    \hat{\mathbf{C}}(\mathbf{r}) = \sum_{i}^{N}T_i(1-\exp(-\sigma_i \delta_i))\mathbf{c}_i,
\end{equation}
where $T_i = \exp(- \sum_{j=1}^{i-1} \sigma_i \delta_i)$, and $\delta_i$ is the distance between adjacent samples. 

\noindent \para{Tensors and low-rank Tensors.} The Plenoxels directly represents a 3D scene by an explicit grid (sparse Tensors)~\citep{fridovich2022plenoxels}. Each grid point stores density and spherical harmonic (SH) coefficients. The color $\mathbf{c}$ is obtained according to the SH and the view direction $\mathbf{d}$ as follows:
\begin{equation}  \label{eq-sh}
\mathbf{c} = S\left(\sum_{\ell=0}^{\ell_{\max }} \sum_{m=-\ell}^{\ell} k_{\ell}^{m} Y_{\ell}^{m}(\mathbf{d})\right),
\end{equation}
where $S: x \mapsto(1+\exp (-x))^{-1}$, $\mathbf{k}=\left(k_{\ell}^{m}\right)_{\ell: 0 \leq \ell \leq \ell_{\max }}^{m:-\ell \leq m \leq \ell}$,  and $k_{\ell}^{m}$ is a set of coefficients, and $l$ is the degree of the SH function $Y_{\ell}^{m}$.

The performance of explicit sparse Tensors depends excessively on the spatial resolution of the grid. In order to reduce the memory footprint caused by the enormous size of the tensor, The VM (Vector-Matrix) decomposition~\citep{chen2022tensorf} factorizes the huge tensor $\mathcal{T} \in \mathbb{R}^{I \times J \times K}$ into low-rank matrices $\mathbf{M}$ and vectors $\mathbf{v}$ as follows:
\begin{equation}  \label{eq-vm}
\mathcal{T}=\sum_{r=1}^{R_{1}} \mathbf{v}_{r}^{1} \circ \mathbf{M}_{r}^{2,3}+\sum_{r=1}^{R_{2}} \mathbf{v}_{r}^{2} \circ \mathbf{M}_{r}^{1,3}+\sum_{r=1}^{R_{3}} \mathbf{v}_{r}^{3} \circ \mathbf{M}_{r}^{1,2},
\end{equation}
where $\mathbf{v}_{r}^{1} \in \mathbb{R}^{I}$, $\mathbf{v}_{r}^{2} \in \mathbb{R}^{J}$, $\mathbf{v}_{r}^{3} \in \mathbb{R}^{K}$, $\mathbf{M}_{r}^{2,3} \in \mathbb{R}^{J \times K}$, $\mathbf{M}_{r}^{1,3} \in \mathbb{R}^{I \times K}$, and $\mathbf{M}_{r}^{1,2} \in \mathbb{R}^{I \times J}$. And $\circ$ represents the outer product. Unlike Plenoxels, VM decomposition does not store color directly but rather features that can be decoded by an MLP.

\noindent \para{Multi-resolution Hash Encoding.} INGP~\citep{muller2022instant} maps a series of grids of different scales to the corresponding feature vectors with a fixed size. It uses a hash function as in Eq.~(\ref{eq-hash}) to map a spatial point in the grid to a hashtable with different resolutions that are adopted to details of different levels of these grids:
\vspace{-1ex}
\begin{equation}  \label{eq-hash}
h(\mathbf{x})=\left(\bigoplus_{i=1}^{d} x_{i} \pi_{i}\right) \quad \bmod \quad S,
\end{equation}
where $\bigoplus$ denotes bit-wise XOR operation. $\pi_{i}$ is an unique large prime number. And $S$ is the hashtable size. These Hashtables store learnable parameters, which are fed to a shallow MLP to interpret densities and colors. INGP effectively reduces the model size by these Hashtables and improves the synthesis quality by introducing multi-resolution.

\begin{table}[t]
\centering
\caption{\textbf{Division of each typical model under our unified two-level view}. Each architecture is represented as a cascade of two modules $\phi_{*}^1$ and $\phi_{*}^2$.}
\begin{tabular}{ccccc}
\hline
methods & $\phi_{*}^1$ & $\phi_{*}^2$ \\ \hline
NeRF   & first K layers & remaining MLP \\
Plenoxels & full & identity function \\
TensoRF & decomposed tensors & MLP decoder \\
INGP & Hashtables & MLP decoder \\ \hline
\end{tabular}
\label{tab:nerf-split}
\end{table}

\subsection{PVD-AL: Progressive Volume Distillation with Active Learning} \label{sec:pvd-al}
We propose PVD-AL to realize the mutual conversion between different architectures. To speed up the training process, we develop a volume-aligned loss and construct a block-wise distillation method based on a unified perspective of various NeRF architectures in PVD-AL. 
We also employ a special treatment of the dynamic density volume range by clipping for its specific numerical instability, which improves the training stability and significantly improves the synthesis quality.
Furthermore, leveraging the unique characteristics of such distillation in NeRF, we devise three levels of active learning strategies that notably enhance distillation performance, surpassing even training a model from scratch. The overview pipeline of PVD-AL is shown in Fig.~\ref{fig:pvd}

\subsubsection{Loss Design} \label{subsec:loss_design}
In our method, we not only use the RGB but also the density, color and an additional intermediate volume feature to calculate loss between different structures. We have observed that the implicit and explicit structures in the hybrid representation are naturally separated and correspond to different learning objectives. Therefore, we consider splitting a model into these similar expression forms so that different parts can be aligned during distillation. Specifically, given a model $\phi_*$, we represent them as a cascade of two modules as follows:
 \begin{equation}
    \phi_*(\mathbf{x}, \mathbf{d}) = \phi_{*}^2(\phi_{*}^1(\mathbf{x}, \mathbf{d})),
 \end{equation}
where * can be either a teacher or a student. We choose the first K layers of MLP in NeRF as $\phi_{*}^1$, and the remaining parts as $\phi_{*}^2$. For hybrid representations, we directly regard the explicit part as $\phi_{*}^1$, and the implicit part as $\phi_{*}^2$. 
As for the purely explicit representation of Plenoxels, we still formulate it into two parts by letting $\phi_{*}^2$ be the identity, though it can be transformed without splitting. The specific splitting of the model is shown in Table~\ref{tab:nerf-split}. Based on the splitting, we design volume-aligned losses as follows:
 \begin{equation}
    \mathcal{L}_{2}^{v} = \left \| \phi_{t}^1(\mathbf{x}, \mathbf{d}) -\phi_{s}^1(\mathbf{x}, \mathbf{d}) \right \| _{2},
\end{equation}
 where $\mathcal{L}_{2}$ indicates the mean-squared error (MSE). The reason for designing this loss is that models in different forms can be mapped to the same space that represents the scene. Our experiments have shown that this volume-aligned loss can accelerate the distillation and improve the quality significantly. The complete loss function during distillation is as follows:
\begin{equation}
\mathcal{L}= \omega_{1}\mathcal{L}_{2}^{v} + \omega_{2}\mathcal{L}_{2}^{\sigma}  + \omega_{3}\mathcal{L}_{2}^{c} + \omega_{4}\mathcal{L}_{2}^{rgb} + \omega_{5}\mathcal{L}_{reg},
\label{equ:loss}
\end{equation}
where $\mathcal{L}^\sigma,\mathcal{L}^c$, and $\mathcal{L}^{rgb}$ denote the density, color and RGB losses, respectively. The last item $\mathcal{L}_{r e g}$ represents the regularization term, which depends on the form of the student model. For Plenoxels and VM-decomposition, we add L1 sparsity loss and total variation (TV) regularization loss.
It should be noted that we only perform density, color, RGB and regularization loss on Plenoxels for its explicit representation.

\begin{algorithm}[!t]
   \caption{PyTorch pseudocode of active learning strategy in PVD-AL.}
   \label{algo:PVD-AL}
    \definecolor{codeblue}{rgb}{0.25,0.5,0.5}
    \lstset{
      basicstyle=\fontsize{7.2pt}{7.2pt}\ttfamily\bfseries,
      commentstyle=\fontsize{7.2pt}{7.2pt}\color{codeblue},
      keywordstyle=\fontsize{7.2pt}{7.2pt},
    }
\begin{lstlisting}[language=python]
# TCP: important camera poses
# TSR: important sample rays
# TSP: important sample points
initial(TCP, TSR, TSP)
for each epoch:
    #generate random poses and sample important poses
    train_poses=[GenRandomPoses(),RandomSelect(TCP)]
    all_batch_rays=SampleRays(train_poses)
    
    for each batch_rays:
        #generate random rays and sample important rays
        train_rays=[batch_rays, RandomSelect(TSR)]
        train_data=SamplePoints(train_rays)
        
        #distillation process
        tea_out=tea_model(train_data)
        stu_out=stu_model(train_data)
        
        #select important points TSP by its weights:PW
        PW=[tea_out[PW], stu_out[PW]]
        sortedPW= GetPointsIdx(sort(PW))
        TSP=sortedPW[:Nsp]
        
        #use TSP to calculate loss sigma and loss color
        loss_s=L2(tea_out[`s'][TSP],stu_out[`s'][TSP])
        loss_c=L2(tea_out[`c'][TSP],stu_out[`c'][TSP])
        
        #calculate loss rgb and other losses
        loss_rgb=L2(tea_out[`rgb'],stu_out[`rgb'])
        loss_other=CalculateOtherLosses()
        
        #select important rays TSR by loss rgb
        sortedRays=GetRaysIdx(sort(loss_rgb))
        TSR[replace_idx]=sortedRays[:Nsr]
        
        #record img loss by train_rays
        loss_img[RaysMapToImgIdx(batch_rays)]+=loss_rgb
        backward_option()
    
    # select important poses TCP by loss img
    sortedImgs=GetPoses(loss_img)
    TCP[repalce_idx]=sortedImgs[:Ncp]

\end{lstlisting}
\vspace{-2 ex}
\end{algorithm}

\subsubsection{Density Range Constrain} \label{subsec:density_range}
We have found that the density loss $\mathcal{L}^\sigma_2$ is hardly directly optimized. We impute this problem to its specific numerical instability. That is, the density reflects the light transmittance of a point in space. When $\sigma$ is greater than or less than a certain value, its physical meaning is consistent (i.e., completely transparent or completely opaque). 
Thus, while the value range of $\sigma$ may be extensive, only a specific interval of density values is crucial. To address this, we constrain the numerical range of $\sigma$ to [$a, b$], and compute $\mathcal{L}_{2}^{\sigma}$ as follows:
\vspace{-2ex}
\begin{equation}
\mathcal{L}_{2}^{\sigma} = \left \| \min (\max (\sigma_{t}, a), b) - \min (\max (\sigma_{s}, a), b) \right \|_2,
\end{equation}
where $\sigma_t$ for teacher model and $\sigma_s$ for student model.
This restriction has an inappreciable impact on the performance of teachers and brings a tremendous benefit to the distillation. Another way to optimize  $\sigma$ involves directly performing the density loss on the $\exp(-\sigma_{i}\delta_{i})$, but we discover it is inefficient since the gradient of $\exp$ is easier to saturate, and it requires computing an exponent that increases the amount of calculation when the block-wise is implemented.

\subsubsection{Block-wise Distillation} \label{subsec:block_wise_diss}
During volume rendering, a significant portion of computation is dedicated to MLP forwarding for each sampled point and integrating the output along each ray. This intensive process notably hampers training and distillation efficiency. In our PVD-AL framework, the utilization of $\mathcal{L}_{2}^{v}$ allows for the implementation of a block-wise strategy to mitigate this issue. Specifically, we initiate training by only forwarding stage 1, followed by sequential execution of stages 2 and 3, as illustrated in Fig.~\ref{fig:pvd}. Consequently, both the student and the teacher networks are spared from the need to forward the entire network and render RGB outputs in the early training phases, significantly accelerating the distillation process.

\subsubsection{Three-Level of Active Learning} \label{subsec:active_learning}
For the purpose of facilitating the most efficient possible transfer of knowledge from teachers to students, we propose an active learning technique in PVD-AL. We continually analyze the camera poses, sample rays, and sample points that are tough to suit for students from coarse to fine, which helps students actively boost their understanding of this crucial knowledge as depicted in Fig.~\ref{fig:pvd}. The specific execution process of the active learning strategy is outlined in Algorithm \ref{algo:PVD-AL}.

\noindent \para{Important camera poses}. Once a teacher model is trained, it can generate images corresponding to any specified camera pose, obviating the need for real data during distill the student model, which is instead guided by the teacher-rendered images. Our experimental findings reveal a significant performance disparity between the teacher and student models under certain poses. To bridge this gap, we propose actively incorporating these challenging poses into the student's distillation process.
Given several camera poses, the images rendered by the teacher and student are respectively marked as $I_{t}$ and $I_{s}$, and the important camera poses $TCP$ can be obtained by the selection function $f_{cp}$:
 \begin{equation}
    TCP = f_{cp}({I}_{t}, {I}_{s}, N_{cp}).
\end{equation}
The function $f_{cp}$ will select the $N_{cp}$ poses with the largest gap between ${I}_{t}$ and ${I}_{s}$.

\noindent \para{Important sample rays}. Sample rays carry varying amounts of information depending on the location they travel through. For example, some rays traverse fully opaque object surfaces, while others may pass through translucent sections. Each type of ray presents a unique challenge to students. Therefore, we provide students with feedback for the rays where there are substantial performance gaps between students and teachers.
Given several sample rays, the RGB values rendered by the teacher and student are respectively marked as $RGB_{t}$ and $RGB_{s}$, and the important sample rays $TSR$ can be obtained by the selection function $f_{sr}$:
 \begin{equation}
    TSR = f_{sr}({RGB}_{t}, {RGB}_{s}, N_{sr}).
\end{equation}
The function $f_{sr}$ will select the $N_{sr}$ rays with the largest gap between ${RGB}_{t}$ and ${RGB}_{s}$.

\noindent \para{Important sample points}. When performing the sigma and color loss as defined in Eq.~(\ref{equ:loss}), by default, all sampling points of the rays are considered. However, not all sampling points contribute equally to the final RGB value. Points closer to the object's surface are more influential, whereas those farther away contribute minimally or nothing at all. Therefore, when distillation, we force the student to prioritize sample points that significantly affect the final RGB value.
Given several sample points along a ray, the weight of each point calculated by teacher and student is respectively marked as $W_{t}$ and $W_{s}$, the important sample points $TSP$ can be obtained by the selection function $f_{sp}$:
 \begin{equation}
    TSP = f_{sp}([{W}_{t}: W_{s}], N_{sp}),
\end{equation}
where $[:]$ means concatenate, and $W_*$ is calculated from Eq.~(\ref{neural_rendering_equation}) as
\begin{equation}  \label{weights}
    W_* = \sum_{i}^{N}T_i(1-\exp(-\sigma_i \delta_i)).
\end{equation}
The function $f_{sr}$ will select the $N_{sp}$ points with the largest value of $W_t$ and $W_s$.


\begin{figure*}[!t]
\centering
\includegraphics[width=0.84\textwidth]{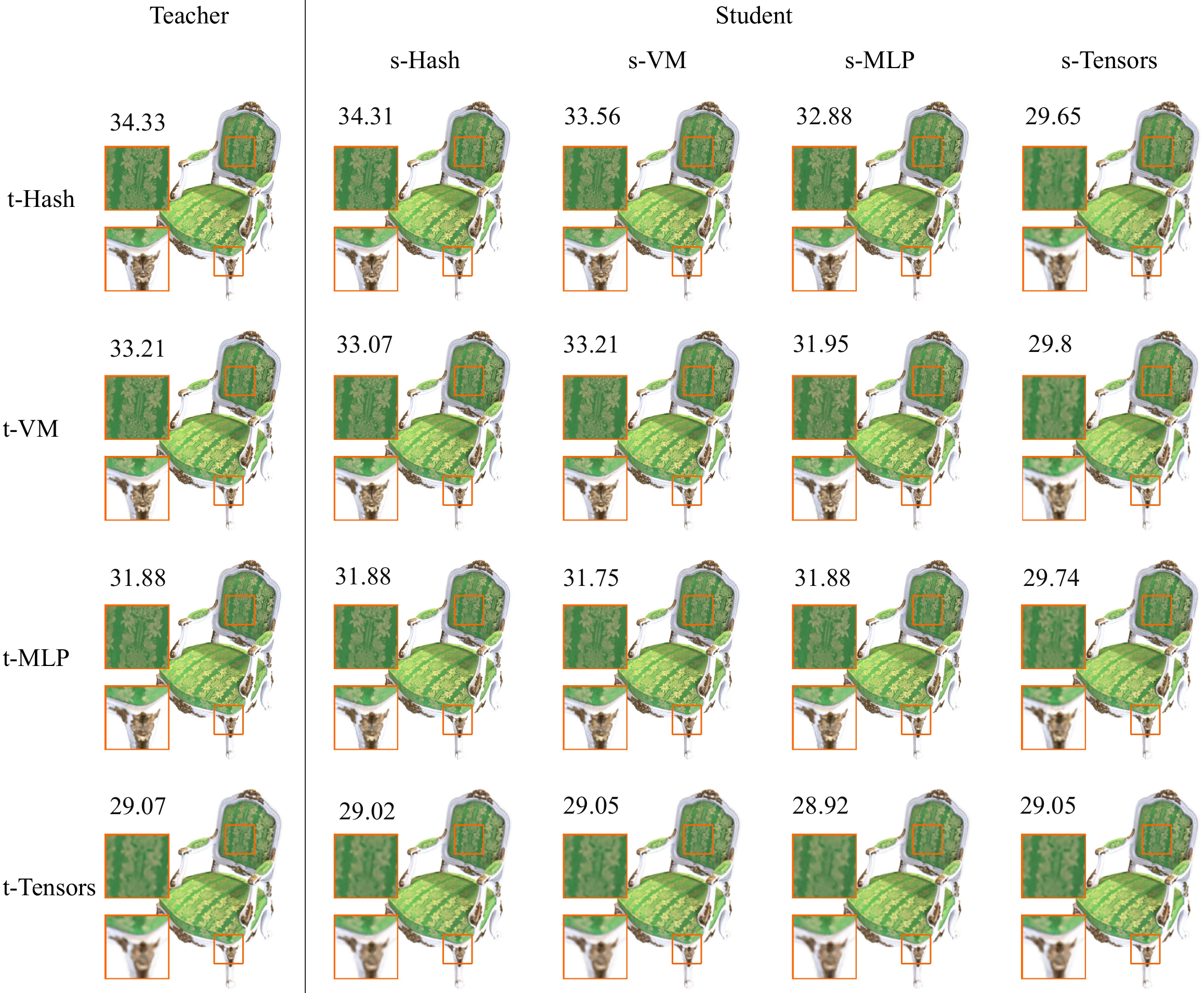} %
\caption{\textbf{Results of mutual-conversion between different NeRF architectures.} We perform the 
PVD-AL between Hashtables, VM-decomposition, MLP and sparse Tensors on the chair scene from the Synthetic-NeRF dataset. The numbers above the image indicate PSNR. The `s-' stands for student, and the `t-' denotes teacher.}
\label{fig:mutual}
\end{figure*}

\begin{table*}[!htb]
\setlength\tabcolsep{3pt}
\centering
\caption{\textbf{Quantitative results of mutual-conversion between different NeRF architectures.}
Metrics are averaged over the Synthetic-NeRF, LLFF, and TanksAndTemple datasets. 
We first train four teachers with different structures for each scene in the three datasets, yielding 84 teacher models in total. We then perform the mutual conversion between four structures for each scene, for a total of 336 distillation experiments.}
\resizebox{\linewidth}{!}{
\input{tables/mutual.tex}

}
\label{tab:mutual}%
\end{table*}

\section{Experiments} \label{sec:experiments}
\begin{figure*}[!hbt]
\centering
\includegraphics[width=1\textwidth]{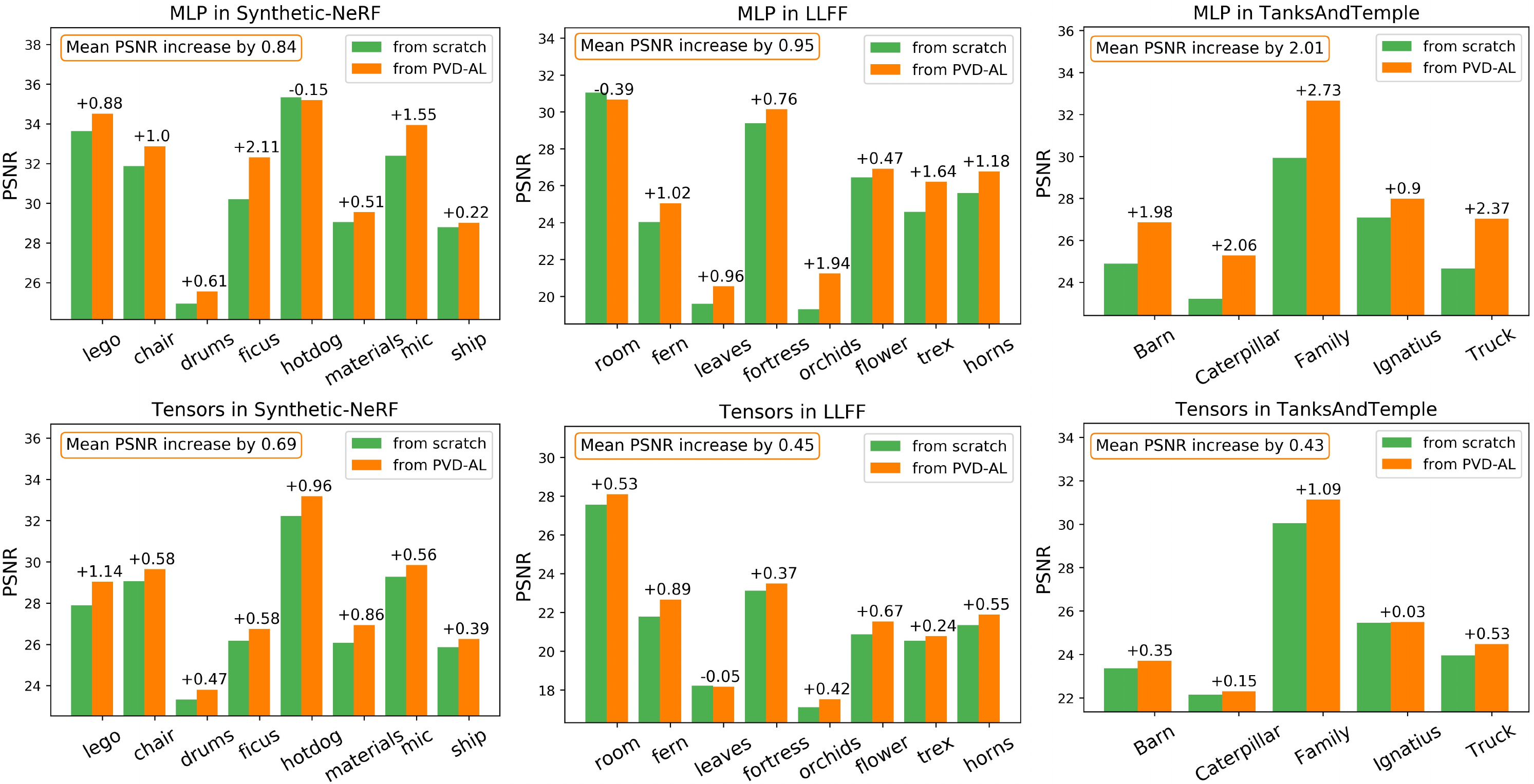}
\caption{\textbf{Quantitative comparison between the model obtained by PVD-AL and that trained from scratch}. PVD-AL demonstrates markedly superior performance in terms of PSNR compared to training a model from scratch.}
\label{fig:psnr-increase}
\end{figure*}

\begin{figure*}[!h]
\centering
\includegraphics[width=0.85\textwidth]{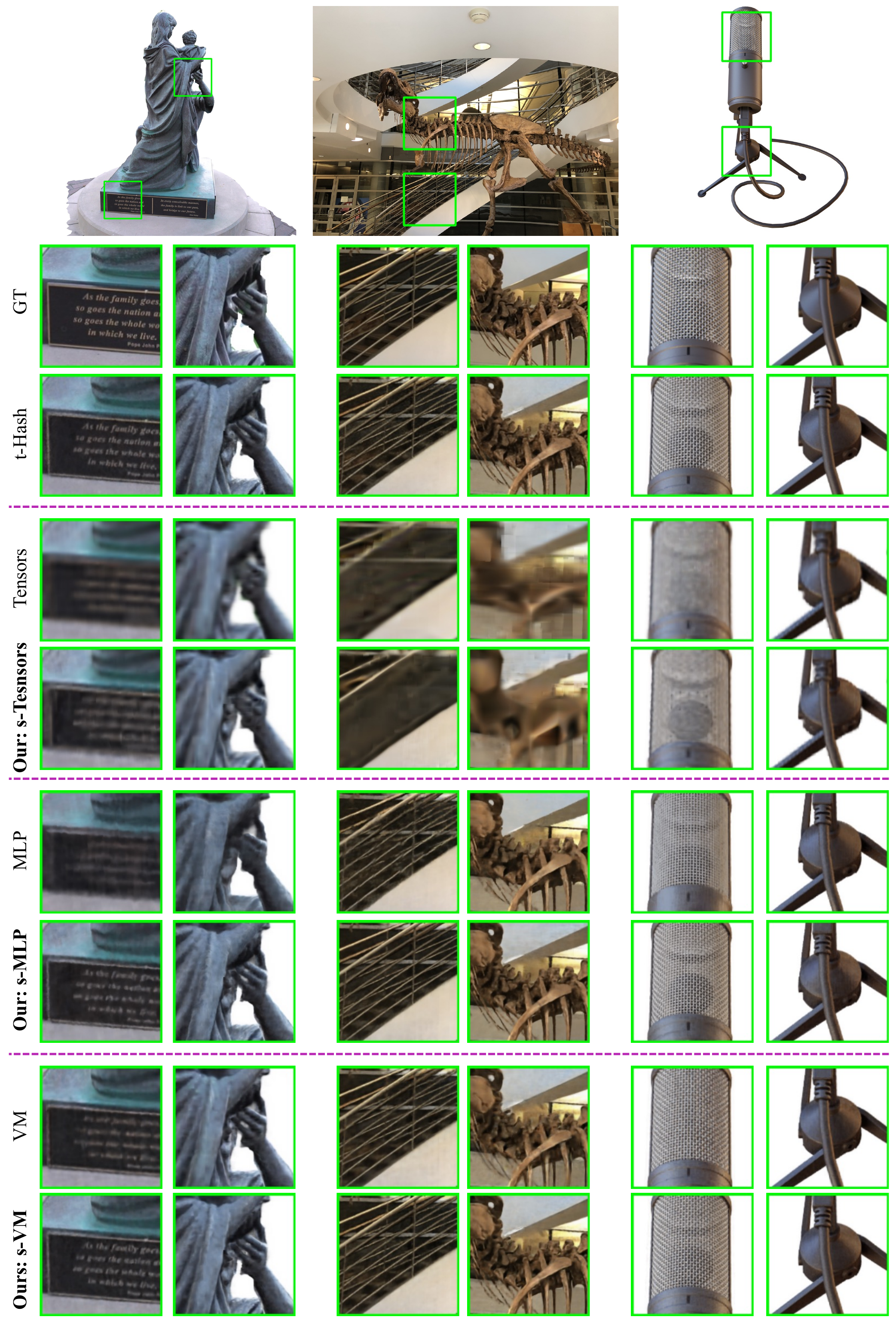} %
\caption{\textbf{Qualitative comparison between the model obtained by PVD-AL and that trained from scratch}. The synthetic quality of our method is preferable to train a model from scratch.}
\label{fig:compare_vis}
\vspace{-2 ex}
\end{figure*}

In this section, we first demonstrate the effectiveness of the PVD-AL method in achieving seamless conversions between different NeRF architectures (Fig.~\ref{fig:mutual} and Table~\ref{tab:mutual}). We prove its significant advantage in enhancing model performance, whereby utilizing PVD-AL as a training tool yields student models that outperform those trained from scratch (Fig.~\ref{fig:shocking}, \ref{fig:psnr-increase} and \ref{fig:compare_vis}).
Subsequently, we analyze the role of the proposed active learning strategies (Fig.~\ref{fig:minmaxmean}) and visualize the important camera poses, sampled rays, and sampled points involved (Fig.~\ref{fig:hard-datas}). We then explore the flexibility of these strategies, applying them as plugins to various methods and significantly improving their performance (Table~\ref{tab:plug} and Fig.~\ref{fig:shocking}).
We then validate the enormous potential of PVD-AL in attribute transfer between different models: By employing PVD-AL, the editing capabilities of different models can be concentrated onto any desired model (Fig.~\ref{fig:shocking} and \ref{fig:edit}). Additionally, we confirm the role of PVD-AL in compressing model parameters and reducing training time (Table~\ref{tab:memory}).
We further apply the capabilities of PVD-AL to terminal interactive tasks, successfully achieving richer scenes and faster interactive models on mobile devices (Fig.~\ref{fig:mobile}), demonstrating the significant application value of PVD-AL.
Finally, we test the results of continued fine-tuning of the distilled models (Table~\ref{tab:finetune}) and conduct ablation experiments (Tables~\ref{tab:ablation} and \ref{tab:division_mlp}) to verify the importance of each component proposed in our method.

\subsection{Implementation Details}

\para{Datasets}. Our experiments are conducted mainly on the three datasets: Synthetic-NeRF ~\citep{mildenhall2020nerf}, forward-facing (LLFF)~\citep{mildenhall2019llff} and TanksAndTemple~\citep{knapitsch2017tanks}. We solely use the aforementioned datasets to train teacher models. 
During the distillation phase, we generate pseudo-labels using the teacher to create synthetic images, without touching the real training data.

\noindent \para{Network Architecture.} We try to stick as close to the original paper settings as feasible for each structure (MLP / sparse Tensors / low-rank Tensors / Hashtables). For MLP~\citep{nerftorch}, positional encoding is also utilized for coordinates and view directions. For sparse Tensors~\citep{fridovich2022plenoxels}, we use spherical harmonics of degree 2, and the $128\times128\times128$ grid for Synthetic-NeRF dataset and TankAndTemple dataset, $512\times512\times128$ grid for LLFF dataset. For low-rank Tensors (VM-decomposition)~\citep{chen2022tensorf}, we take 48 components in total. For Hashtables~\citep{muller2022instant}, we set the coarsest resolution, the finest resolution, levels, hashtable size and feature dimensions to 16, $2048\times\text{scene scope}$, 14, $2^{19}$, and 2 respectively. 

\noindent \para{Training and Distilling Details.} We implement our method with the PyTorch framework~\citep{paszke2019pytorch} to train teachers and distill students. We use Adam optimizer~\citep{kingma2014adam} with initial learning rates of 0.02 and run 20k steps with a batch size of 4096 rays. For distilling, we initial the loss rate for volume-aligned, density, color, and RGB with 2e-3, 2e-3, 2e-3, and 1, respectively. The first stage consumes 3k steps, the second stage consumes 5k steps, and the third stage will take all the resting steps.
Three levels of active learning strategy will be incorporated into the rest training process after the first two phases have been completed. We set 10\% of important rays as important rays and 30\% sample points as important points in each training iteration, and 10\% camera poses as important poses in each training epoch.
All the experiments are performed on a single NVIDIA V100 GPU. 

\subsection{Mutul-Conversion between Different Architectures}
\subsubsection{Quantitative Results}
We first train four teachers with different structures for each scene in the three benchmark datasets, yielding 84 teacher models in total, and then perform mutual conversion between the four structures for each scene, totaling 336 distillation experiments. We then average the PSNR / SSIM (higher is better) and LPIPS~\citep{zhang2018lpips} (lower is better) results for these distillation experiments, which are shown in Table~\ref{tab:mutual}. This table demonstrates that our method is able to successfully complete the mutual-conversion between four different structures, and the students who are the outcome of our method display exceptionally competitive performance.

We further experimentally validated the assistance provided by PVD-AL in enhancing the performance limit of the student, wherein a student distilled from a high-performance teacher exhibits better performance than one trained from scratch, as seen in Fig.~\ref{fig:psnr-increase}.
For instance, when distilling a Hashtables-based model to an MLP-based model, the PSNR on the TanksAndTemple dataset increases by 2dB compared to training the MLP-based model from scratch, demonstrating the superiority of our distillation scheme.

Further analysis in Table~\ref{tab:mutual} reveals that the student's performance is constrained by two factors: the teacher's performance and the student's own capacity. When the student's modeling abilities are inferior to or comparable to those of the teacher, the student can strive to mimic the teacher's performance to the fullest extent possible. Conversely, if a student's modeling capabilities surpass those of the teacher, the student can undergo further enhancement through fine-tuning, which will be covered in detail in Subsection~\ref{subsec:finetune}.

\begin{figure*}[!hbt]
\vspace{+1 ex}
  \centering
  \includegraphics[width=1.\textwidth]{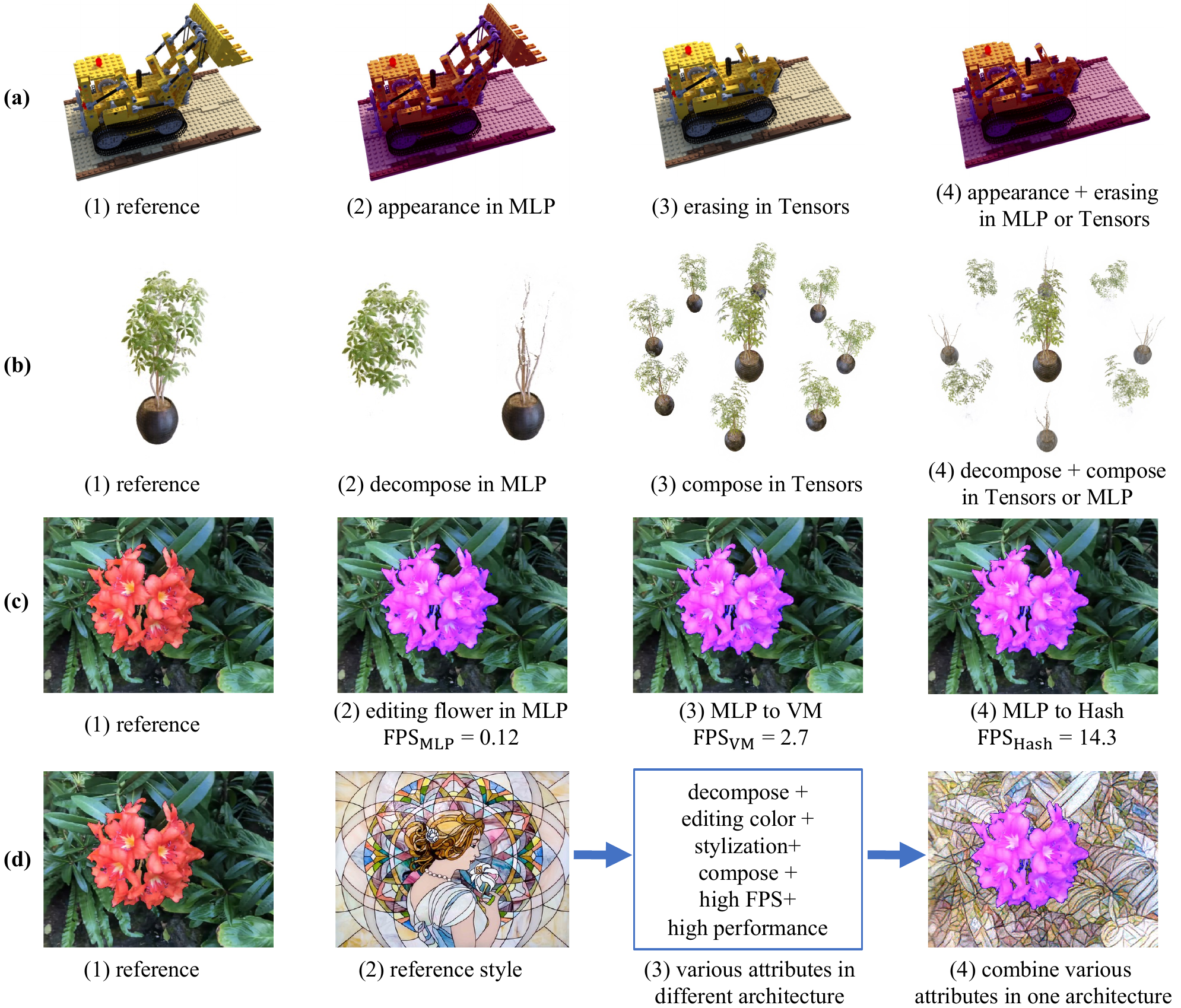}
  \caption{\textbf{Transferring of different characteristics between diverse architectures}. PVD-AL enables the superposition of distinct editing skills, high performance, and high FPS benefits. This is difficult to accomplish previously on a single structure.}
  \label{fig:edit}
  \end{figure*}

\begin{figure*}[!t]
\centering
\includegraphics[width=1\textwidth]{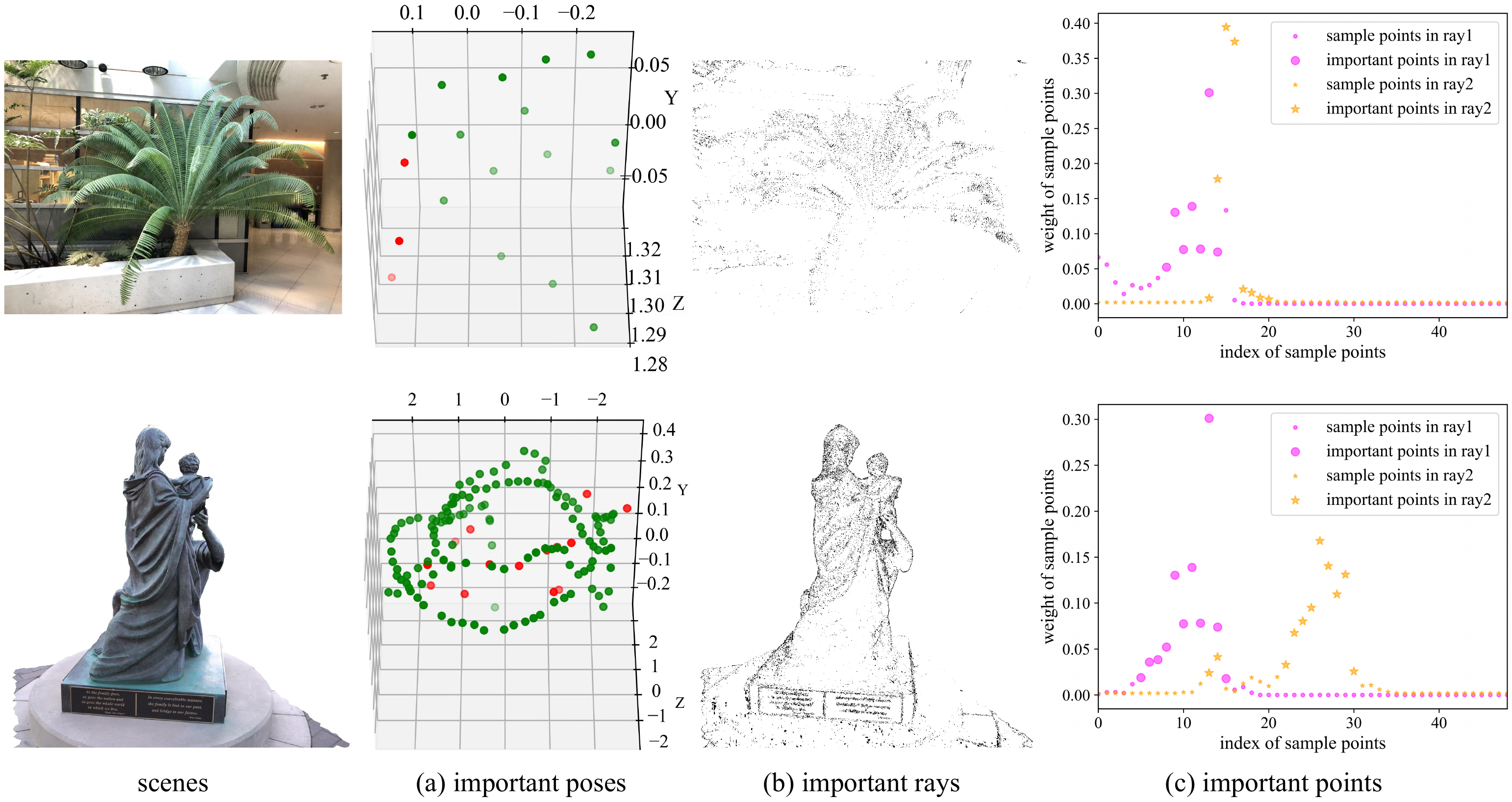}
\caption{\textbf{Visualization of important camera poses, sample rays, and sample points during the distillation process}. (a) illustrates the distribution of camera poses, where red indicates important poses. (b) demonstrates the distribution of pixels corresponding to important rays. (c) showcases the sample points along rays and their corresponding weights, where larger points denote the important sample points.}
\label{fig:hard-datas}
\end{figure*}

\begin{figure}[ht]
  \centering
  \includegraphics[width=0.48\textwidth]{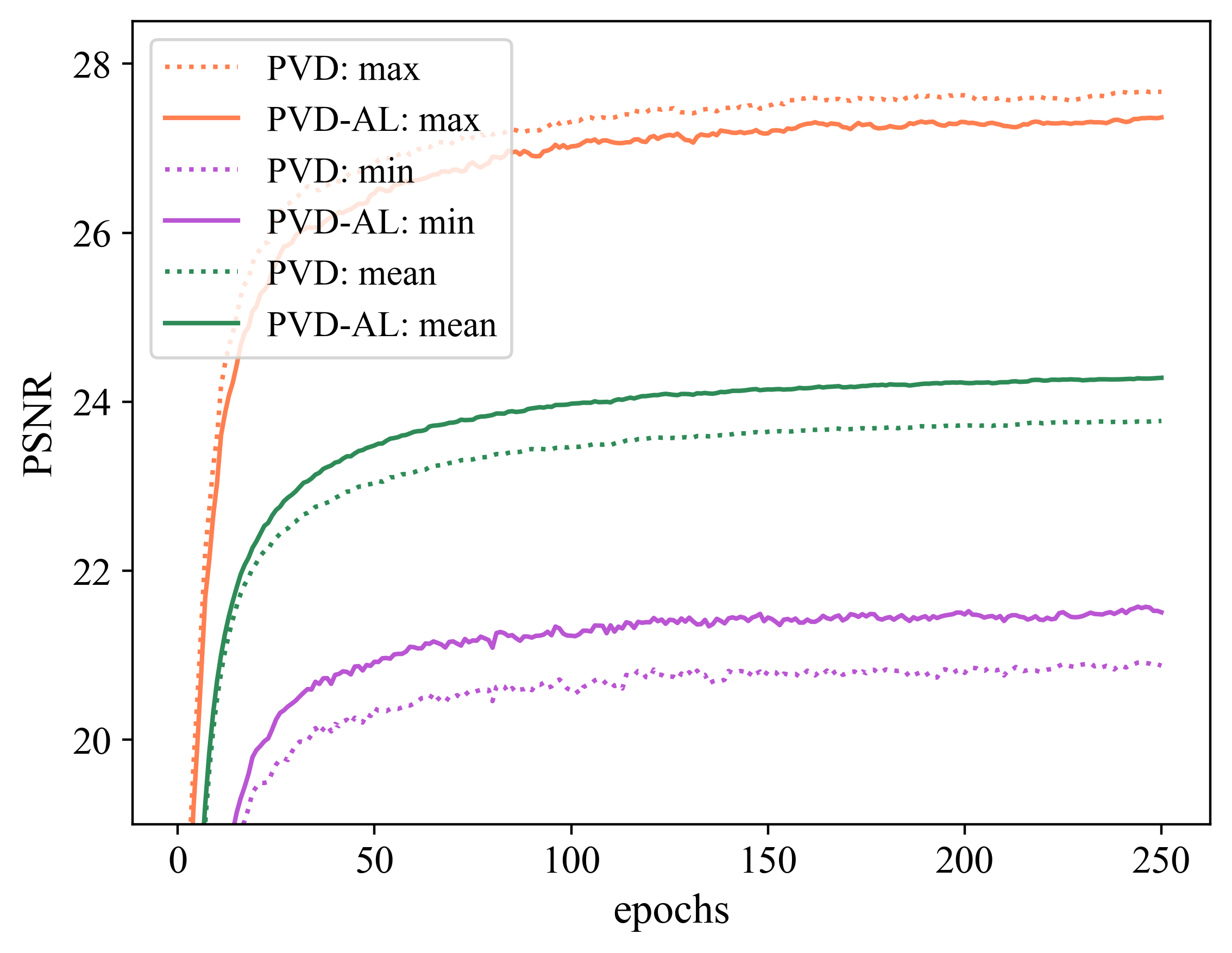} 
  \caption{\textbf{Trend of maximum, minimum and average PSNR on validation set when distilling VM to Tensors in Truck scene from TanksAndTemples dataset}. Solid lines indicate the utilization of active learning, while dashed lines represent its absence.}
  \label{fig:minmaxmean}
  \end{figure}

\subsubsection{Qualitative Results}  
Fig.~\ref{fig:mutual} illustrates an example of mutual-conversion visualization results on the chair scene from the Synthetic-NeRF dataset amongst different architectures of Hashtables, VM-decomposition, MLP, and sparse Tensors. 
The visual quality of the student is frequently indistinguishable from that of the teacher, illustrating the outstanding features of our method for sustaining synthesis quality.
We also present a visual comparison between the model obtained by our method and the model trained from scratch, as shown in Fig.~\ref{fig:compare_vis}.
It can be observed that the rendering images produced by PVD-AL are superior to those produced by training the student from scratch. This advancement is mostly attributable to our distillation method across distinct structures, which makes it possible for an experienced teacher to break the upper limit of a student's capabilities. Therefore, PVD-AL can serve as a novel training tool to attain higher-performing NeRF series models.

\subsection{Editing Ability Conversion} \label{subsce:edit}
In addition to the various advantages mentioned above, one of the most important features of our approach is the implementation of property migration between different structures, which is the key motivation of this paper, i.e., \textit{there is no single ``best" architecture for neural rendering}. 
For example, we find that Hashtables-based architecture is fast and produces high-quality modeling but lacks clear geometric interpretation due to the spatial aliasing caused by its Hashtables. Consequently, tasks like geometric editing, such as erasing or combining, are challenging compared to those facilitated by explicit Tensor structures.
Similarly, the MLP offers an implicit space conducive to embedding various features, enabling artistic-style rendering. Nonetheless, its geometric structure remains ambiguous.

As one concrete example, we first run an editing experiment by distilling between MLP and Tensors. We train an MLP-based model with appearance~\citep{martin2021nerfinthewild} in Lego scene as shown in Fig.~\ref{fig:edit}\textcolor{blue}{(a)(2)}, a Tensors-based model with erasing bucket as shown in Fig.~\ref{fig:edit}\textcolor{blue}{(a)(3)}. 
Then, we can transfer the appearance from MLP to Tensors by PVD-AL and simply erase the bucket in the scene using Tensors' clear geometric. Finally, we obtain a scene with both appearance and geometry changes represented by Tensors.
Likewise, we can also transfer the erasing ability from Tensors to MLP to empower MLP with two editing capabilities as shown in Fig.~\ref{fig:edit}\textcolor{blue}{(a)(4)}.

The decomposition and composition of scenes is another illustration of the migration property between implicit MLP and explicit Tensors. The hidden space vector of the MLP can be employed as a retrievable feature to achieve semantic-level deconstruction~\citep{kobayashi2022decomposingEditing}, as shown in Fig.~\ref{fig:edit}\textcolor{blue}{(b)(2)}. 
The unambiguous spatial geometric relationship of explicit Tensors can be used to combine various scenes~\citep{tang2022CCNeRF}, as shown in Fig.~\ref{fig:edit}\textcolor{blue}{(b)(3)}. Our method may simultaneously realize decomposition and composition and describe it as any architecture like in Fig.~\ref{fig:edit}\textcolor{blue}{(b)(4)}.

The third illustration is the aesthetic editing of a specific object in a scene utilizing implicit MLP-based models~\citep{kobayashi2022decomposingEditing}, as seen in Fig.~\ref{fig:edit}\textcolor{blue}{(c)(2)}.
Nevertheless, MLP-based models typically suffer from sluggish inference speed and low FPS, reducing their usefulness in practical applications.
In this instance, we can transfer the MLP scenes with artistic effects to other faster architectures (such as VM and Hash) by our PVD-AL, as depicted in Fig.~\ref{fig:edit}\textcolor{blue}{(c)(3)} and Fig.~\ref{fig:edit}\textcolor{blue}{(c)(4)}.

It should be emphasized that our method does not place a cap on the number of characteristics that can be migrated. Hence, all the characteristics shown in Fig.~\ref{fig:edit} can be incorporated into a single model. By transforming the appearance, erasing, editing color, decomposition, composition, high-FPS and high-performance properties from many structures to one structure, we have successfully merged the benefits of various structures, which was difficult to accomplish previously on a single structure, as seen in Fig.~\ref{fig:shocking}\textcolor{blue}{(a)} and Fig.~\ref{fig:edit}\textcolor{blue}{(d)}.

\begin{figure*}[!h]
\centering
\includegraphics[width=0.85\textwidth]{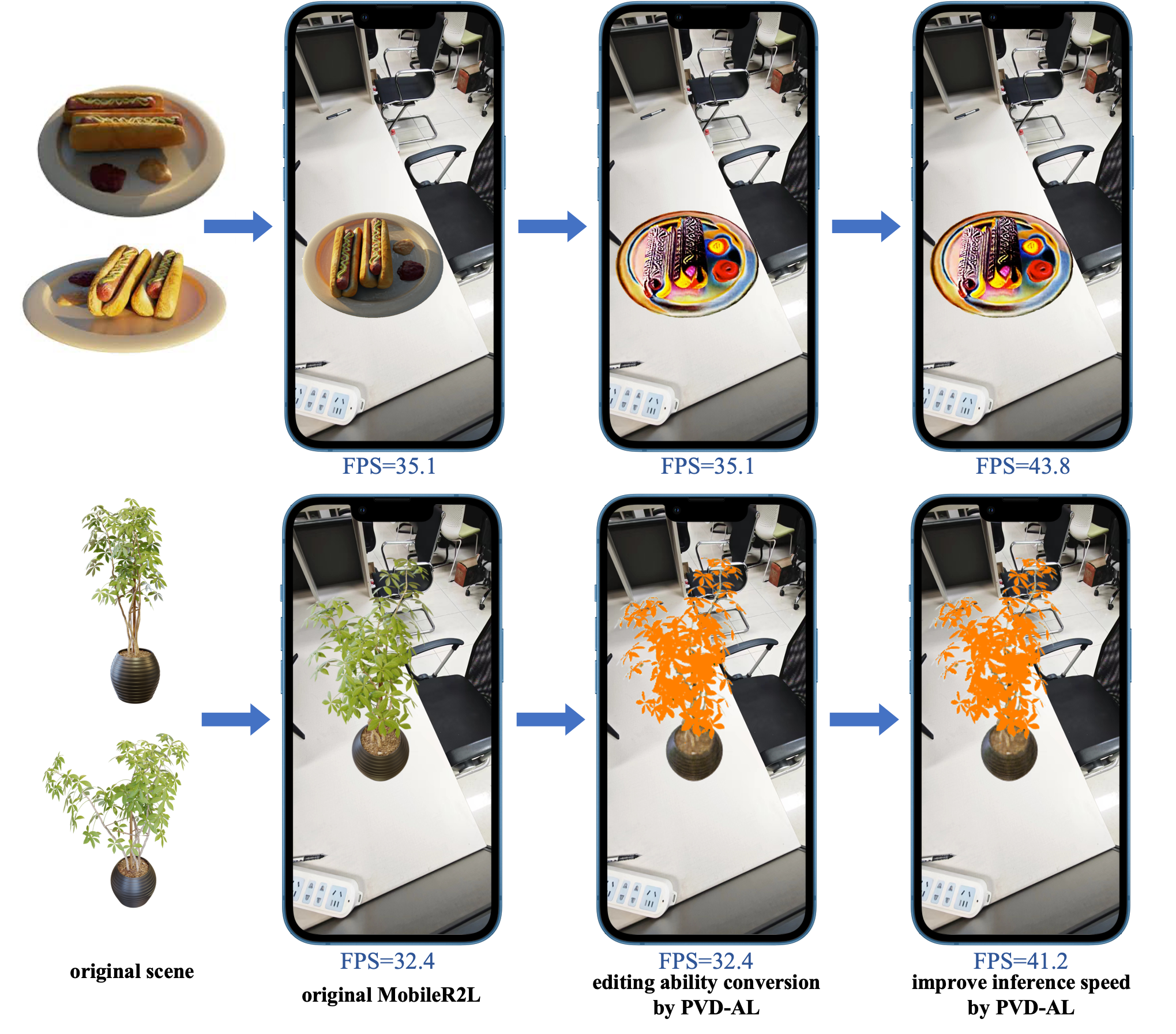}
\caption{\textbf{Expansion of virtual interaction on iPhone 13 device}. By leveraging the editing capabilities and model compression efficiency of PVD-AL, the diversity and real-time responsiveness of interactions can be effectively enhanced, a feat unattainable by the vanilla mobileR2L method~\citep{cao2022mobiler2l}.}
\label{fig:mobile}
\end{figure*}

\subsection{Properties of Active Learning} \label{subsec:result-active-learning}
\subsubsection{Role of Active Learning}
We analyze the PSNR trend for our method with (PVD-AL) and without (PVD) active learning during training. Fig.~\ref{fig:minmaxmean} illustrates the variations in the maximum, minimum, and mean PSNR values across all images in the validation set. The results show that while the maximum PSNR of PVD-AL is slightly lower, the minimum PSNR is significantly higher compared to PVD. This suggests that the active learning technique enhances student performance on challenging samples, leading to a more balanced overall performance and a higher average PSNR.

\subsubsection{Visualization of Active Learning}
We extract the important camera poses, the pixel positions corresponding to important rays, and the weights of sample points found by our active learning strategy and depict them in Fig.~\ref{fig:hard-datas}. 

Fig.~\ref{fig:hard-datas}\textcolor{blue}{(a)} illustrates that important camera poses are concentrated in specific areas rather than randomly distributed. This suggests that students encounter difficulties primarily with a small subset of camera poses, and focusing more on these poses during training can notably enhance the student's performance.

For the display of important rays in Fig.~\ref{fig:hard-datas}\textcolor{blue}{(b)}, it is evident that, during the distillation process, our active learning strategy concentrates mostly on the high-frequency information of the images, which is consistent with the results of Fig.~\ref{fig:compare_vis} that the synthetic quality in high-frequency areas is improved. 
Notably, the absence of important rays in spatially empty zones suggests that students can readily adapt to these regions. Consequently, indiscriminate fitting of such areas would result in the student engaging in futile efforts. Our approach addresses this issue by directing the student's learning focus towards the most crucial regions within a scene, thus optimizing learning potential.

In Fig.~\ref{fig:hard-datas}\textcolor{blue}{(c)}, we present two important rays and illustrate the weights assigned to the sample points along each ray.
These weights exhibit a clear impulse distribution, suggesting that sample points with higher weights have a greater impact on the resulting RGB values. By identifying these crucial points to students in advance, they can better prioritize fitting them during the distillation process.

\begin{table}[t]
\setlength\tabcolsep{3pt}
  \centering
  \caption{\textbf{Scalability of the proposed active learning strategy to other NeRF-based distillation tasks}. 
  ``+AL" indicates that active learning is employed as a plug-in, where the performance is further improved.}
  \resizebox{1.\linewidth}{!}{
    \input{tables/plug.tex}

    }
  \label{tab:plug}
\end{table}

\subsubsection{Improving Other Methods As a Plug-in}
Our active learning strategy not only applies to PVD-AL but is also readily extendable to various NeRF-based distillation tasks, thereby enhancing their performance. For example, we deploy this strategy to augment the R2L method~\citep{wang2022r2l}, which distills a radiation field model into a light field model, as well as KiloNeRF~\citep{reiser2021kiloNeRF}, which distills a high-capacity MLP into thousands of smaller MLP. For ease of implementation, we exclusively focus on important sample rays and camera poses for the aforementioned models. The visualization outcomes are depicted in Fig.~\ref{fig:shocking}\textcolor{blue}{(c)}, while the metric results are presented in Table~\ref{tab:plug}.
It can be seen that the integration of our active learning strategy has markedly enhanced the performance of R2L and KiloNeRF, underscoring the generalizability of our approach.
Besides, it is worth noting that the design of the three levels of the active learning strategy in this study is decoupled and allows users to integrate only certain levels, making it highly adaptable when applied to other methods.
\begin{table}[t]
\setlength\tabcolsep{2pt}
  \centering
  \caption{\textbf{Ability of PVD-AL to compress model parameters and reduce training time}. The teacher is based on the representation of Hashtables.}
  \resizebox{1\linewidth}{!}{
    \input{tables/time_memory.tex}

    }
  \label{tab:memory}
\end{table}

\begin{table*}[]
\setlength\tabcolsep{2.pt}
\centering
\caption{\textbf{Finetuning results of the model obtained from PVD-AL}. Metrics are
averaged over the scenes on the Synthetic-NeRF, LLFF, and TanksAndTemple datasets. Case 1: The teacher is superior in modeling capabilities. Case 2:The student is superior in modeling capabilities.}
  \resizebox{\linewidth}{!}{
    \input{tables/finetune_all_scene.tex}
    }
\label{tab:finetune}
\end{table*}

\subsection{Compressing Model Size and Training Time}
PVD-AL is able to produce better outcomes with a more concise parameter representation for students with inferior modeling capabilities since it can help students study more thoroughly.
In order to confirm this, we compress the parameters of the NeRF and Plenoxels, and compare them trained from scratch with the same number of iterations. The results are displayed in Table~\ref{tab:memory}.
The model produced by PVD-AL still performs fairly or even better with fewer parameters than the model trained from scratch.
Consequently, our method can be utilized as a compression tool, which is advantageous in situations with limited hardware resources.

\subsection{Deployment to Mobile Device} \label{subsec:mobile}
We further explore the application value of PVD-AL. Specifically, we integrate active learning as a plugin into the mobileR2L method~\citep{cao2022mobiler2l} to enhance its rendering quality. Subsequently, we leverage PVD-AL's capability of attribute transfer to generate new scenes with editing effects. Finally, utilizing PVD-AL's ability of model compression, we obtain models with smaller parameter sizes and faster running speeds and deploy these models on iPhone~13 devices as shown in Fig.~\ref{fig:mobile}.
It is evident that PVD-AL successfully extends mobileR2L's functionality on resource-constrained mobile devices, enabling the generation of distinctive, interactively editable 3D scenes and significantly enhancing user interactions' real-time responsiveness.
 
\subsection{Finetuning Effects} \label{subsec:finetune}

The finetuning effects after distillation, as delineated in Table~\ref{tab:finetune}, are categorized into two cases.
Case 1: When the teacher exhibits superior modeling capabilities, finetuning yields minimal benefits. The main reason is that a superior teacher can provide sufficient pseudo datasets to train students adequately. Therefore, the final performance boost from using real data is limited.
Case 2: The student demonstrates superior modeling capabilities, finetuning enhances the student's performance. This improvement occurs because the student’s performance is capped by the teacher when distilling.
It's essential to emphasize that our method primarily aims to leverage the distinct properties of various structures, as elaborated in Subsection~\ref{subsce:edit}. Therefore, it still remains meaningful when distilling a student architecture with inferior modeling capabilities. To mitigate unnecessary information loss, employing common techniques like augmenting model parameters can help better align the capabilities of teachers and students.

\subsection{Ablation Study of PVD-AL Components}
\label{subsec:ablation}

\begin{table}[!t]
  \centering
  \caption{\textbf{Ablation study of each designed component}. Metrics are averaged over the 5 scenes from TanksAndTemples dataset in the conversion from VM-decomposition to MLP.}
  \resizebox{0.999\linewidth}{!}{
\input{tables/ablation.tex}

}
\label{tab:ablation}%
\end{table}

To evaluate the importance of each component of PVD-AL, we conduct several groups of ablation experiments, the results of which are presented in Table~\ref{tab:ablation}. The experiments indicate that our developed loss contributes to a PSNR improvement ranging from approximately 0.4 to 2.2 dB. Furthermore, we observe that unregulated density values adversely affect performance. Additionally, omitting the block-wise technique during distillation led to inferior outcomes. Lastly, we underscore the importance of the active learning strategy at each level.

\begin{table}[t]
  \centering
  \caption{\textbf{Ablation study of the division position for MLP}. Metrics are
averaged over the 8 scenes on Synthetic-NeRF dataset in the conversion from Hashtables to MLP.}
\input{tables/division_mlp.tex}
\label{tab:division_mlp}%
\end{table}
We also conduct an ablation experiment on the K value in Table~\ref{tab:nerf-split}, and the average PSNR and training time are as shown in Table~\ref{tab:division_mlp}. In general,
picking a proper K value is a process of striking a balance between performance and training time.
In general, a larger K results in a higher PSNR, indicating that using more layers to fit Hashtables can enhance performance. Conversely, a smaller K reduces training time due to our blockwise distillation strategy. In this context, K=4 would be a Pareto optimum.

\section{Discussion and Conclusion}
In this work, we present PVD-AL, a systematic distillation method that allows conversions between different NeRF architectures, including MLP, sparse Tensors, low-rank Tensors, and Hashtables, while maintaining high synthesis quality. Empirical experiments solidly demonstrate the efficiency of our approach, on both synthetic and real-world datasets.

Critical to the success of PVD-AL is the meticulous design of loss functions, progressive and blockwise distillation schemes utilizing intermediate volume representations, and special treatment of density values.
Additionally, we introduce a three-level active learning methodology. This strategy continually evaluates and updates the challenging camera poses, sample rays, and sample points for students, enhancing their comprehension of these crucial elements.
This three-level active learning strategy is decoupled and adaptable, serving as versatile plug-ins for other distillation tasks.

By breaking down barriers between different NeRF architectures, PVD-AL can be easily employed to assist downstream tasks. For instance, it serves as a tool for efficiently training and compressing models, acquiring them in a faster, higher-performing manner compared to training from scratch.
PVD-AL also allows for the fusion of various attributes across different structures, enabling the acquisition of models with multiple editing abilities. Furthermore, it facilitates the transformation of models into a specific one that runs more efficiently, meeting the real-time demands of hardware resources (such as mobile devices).

In view of the limitations of PVD-AL, we consider possible corresponding solutions. For instance, the student's performance is inherently bounded by the teacher's capability. This limitation can be addressed either by fine-tuning the student or enhancing the teacher's modeling capacity. Furthermore, concurrent execution of both teacher and student models during distillation may trigger memory overflow issues. In such instances, a viable solution is to execute the distillation process sequentially.

\noindent\para{Data availability statement}
The data utilized in Tables 2 to 7 and Figs. 1 to 9 all can be accessed from the following public repositories:
\begin{itemize}
\item Synthetic-NeRF dataset: \url{https://drive.google.com/drive/folders/128yBriW1IG_3NJ5Rp7APSTZsJqdJdfc1}
\item LLFF dataset: \url{https://cseweb.ucsd.edu/~viscomp/projects/LF/papers/SIG19/lffusion}
\item TanksAndTemples dataset: \url{https://www.tanksandtemples.org/download}
\end{itemize}

\begin{acknowledgements}
This work is supported by the National Natural Science Foundation of China (U20B2042, 62076019).
\end{acknowledgements}

\bibliographystyle{spbasic}      
\bibliography{ref}   %

\end{document}

%% file: tables/mutual.tex
    \begin{tabular}{l|cccc|cccc|cccc|cccc}
    \toprule
    \multirow{4}[4]{*}{} & \multicolumn{4}{c|}{\multirow{2}[1]{*}{PSNR$\uparrow$}} & \multicolumn{4}{c|}{\multirow{2}[1]{*}{SSIM$\uparrow$}} & \multicolumn{4}{c|}{\multirow{2}[1]{*}{$\text{LPIPS}_{alex}\downarrow$}} & \multicolumn{4}{c}{\multirow{2}[1]{*}{$\text{LPIPS}_{vgg}\downarrow$}} \\
          & \multicolumn{4}{c|}{}         & \multicolumn{4}{c|}{}         & \multicolumn{4}{c|}{}         & \multicolumn{4}{c}{} \\
          & t-Hash & t-VM  & t-MLP & t-Tensors & t-Hash & t-VM  & t-MLP & t-Tensors & t-Hash & t-VM  & t-MLP & t-Tensors & t-Hash & t-VM  & t-MLP & t-Tensors \\
\cmidrule{2-17}          & \multicolumn{16}{c}{\textbf{Synthetic-NeRF}} \\
    \midrule
    \multicolumn{1}{r|}{                            teacher} & 32.58 & 31.52 & 30.78 & 27.49 & 0.96  & 0.955 & 0.946 & 0.917 & 0.032 & 0.04  & 0.049 & 0.122 & 0.055 & 0.061 & 0.075 & 0.112 \\
    \midrule
    s-Hash$^{}$ & 32.58 & 31.35 & 30.75 & 27.44 & 0.96  & 0.954 & 0.947 & 0.915 & 0.032 & 0.042 & 0.051 & 0.117 & 0.055 & 0.067 & 0.077 & 0.116 \\
    s-VM$^{}$ & 31.83 & 31.52 & 30.57 & 27.48 & 0.957 & 0.955 & 0.945 & 0.916 & 0.038 & 0.04  & 0.053 & 0.121 & 0.06  & 0.061 & 0.077 & 0.114 \\
    s-MLP$^{}$ & 31.63 & 30.78 & 30.78 & 27.38 & 0.953 & 0.948 & 0.946 & 0.914 & 0.043 & 0.05  & 0.049 & 0.119 & 0.069 & 0.076 & 0.075 & 0.118 \\
    s-Tensors$^{}$ & 28.19 & 28.2  & 28.09 & 27.49 & 0.924 & 0.923 & 0.921 & 0.917 & 0.102 & 0.103 & 0.105 & 0.122 & 0.105 & 0.107 & 0.11  & 0.112 \\
    \midrule
          & \multicolumn{16}{c}{\textbf{LLFF}} \\
    \midrule
    \multicolumn{1}{r|}{                                  teacher} & 26.7  & 25.27 & 25    & 21.33 & 0.832 & 0.777 & 0.748 & 0.59  & 0.13  & 0.196 & 0.227 & 0.53  & 0.231 & 0.295 & 0.344 & 0.512 \\
    \midrule
    s-Hash$^{}$ & 26.68 & 25.36 & 25.09 & 21.39 & 0.83  & 0.782 & 0.754 & 0.592 & 0.133 & 0.195 & 0.233 & 0.526 & 0.23  & 0.287 & 0.334 & 0.511 \\
    s-VM$^{}$ & 25.98 & 25.24 & 24.85 & 21.39 & 0.793 & 0.774 & 0.74  & 0.591 & 0.19  & 0.197 & 0.251 & 0.529 & 0.29  & 0.292 & 0.346 & 0.511 \\
    s-MLP$^{}$ & 25.95 & 24.89 & 25.01 & 21.33 & 0.786 & 0.741 & 0.748 & 0.592 & 0.201 & 0.253 & 0.227 & 0.53  & 0.315 & 0.361 & 0.347 & 0.524 \\
    s-Tensors$^{}$ & 21.87 & 21.26 & 21.34 & 21.44 & 0.611 & 0.59  & 0.586 & 0.591 & 0.512 & 0.543 & 0.536 & 0.527 & 0.499 & 0.514 & 0.517 & 0.512 \\
    \midrule
          & \multicolumn{16}{c}{\textbf{TanksAndTemples}} \\
    \midrule
    \multicolumn{1}{r|}{teacher} & 29.26 & 27.27 & 25.96 & 25    & 0.915 & 0.897 & 0.879 & 0.865 & 0.106 & 0.189 & 0.201 & 0.279 & 0.134 & 0.182 & 0.2   & 0.225 \\
    \midrule
    s-Hash$^{}$ & 29.24 & 27.13 & 25.94 & 24.92 & 0.915 & 0.893 & 0.88  & 0.863 & 0.106 & 0.184 & 0.203 & 0.271 & 0.134 & 0.183 & 0.201 & 0.229 \\
    s-VM$^{}$ & 28.3  & 27.27 & 25.89 & 24.96 & 0.906 & 0.895 & 0.879 & 0.865 & 0.153 & 0.188 & 0.201 & 0.28  & 0.165 & 0.181 & 0.204 & 0.227 \\
    s-MLP$^{}$ & 27.97 & 26.71 & 25.93 & 24.78 & 0.9   & 0.887 & 0.876 & 0.863 & 0.152 & 0.201 & 0.218 & 0.273 & 0.175 & 0.197 & 0.201 & 0.23 \\
    s-Tensors$^{}$ & 25.43 & 25.31 & 24.84 & 24.98 & 0.865 & 0.867 & 0.863 & 0.865 & 0.263 & 0.262 & 0.266 & 0.281 & 0.228 & 0.222 & 0.224 & 0.225 \\
    \bottomrule
    \end{tabular}%

%% file: tables/plug.tex

    \begin{tabular}{l|cc|cc|cc}
    \toprule
          & \multicolumn{2}{c|}{Synthetic-NeRF} & \multicolumn{2}{c|}{LLFF} & \multicolumn{2}{c}{TanksAndTemples} \\
    \midrule
          & PSNR  & SSIM  & PSNR  & SSIM  & PSNR  & SSIM \\
    \midrule
    teacher & 30.42 & 0.946 & 27.54 & 0.896 & -     & - \\
    R2L & 30.02 & 0.938 & 27.19 & 0.89  & -     & - \\
    R2L+AL & \textbf{30.35} & \textbf{0.94} & \textbf{27.27} & \textbf{0.891} & -     & - \\
    \midrule
    teacher & 31.14 & 0.952 & -     & -     & 28.26 & 0.901 \\
    KiloNeRF & 28.02 & 0.921 & -     & -     & 26.18 & 0.793 \\
    KiloNeRF+AL & \textbf{29.21} & \textbf{0.943} & -     & -     & \textbf{27.41} & \textbf{0.847} \\
    \bottomrule
    \end{tabular}%

%% file: tables/time_memory.tex
    \begin{tabular}{l|ccc|ccc|ccc}
    \toprule
          & \multicolumn{3}{c|}{Synthetic-NeRF} & \multicolumn{3}{c|}{LLFF} & \multicolumn{3}{c}{TanksAndTemples} \\
\cmidrule{2-10}    \multicolumn{1}{c|}{Method} & PSNR  & Memory & Time  & PSNR  & Memory & Time  & PSNR  & Memory & Time \\
    \midrule
    Plenoxles & 27.74 & 560M  & \textbf{0.34h} & 21.72 & 2.2G  & \textbf{0.76h} & \textbf{25.75} & 4.4G  & \textbf{0.62h} \\
    s-Tensors & \textbf{28.19} & \textbf{233M} & 0.42h & \textbf{21.87} & \textbf{1.8G} & 0.91h & 25.43 & \textbf{1.8G} & 0.78h \\
    \midrule
    NeRF  & \textbf{31.21} & 14M   & 19.5h & 25.2  & 14M   & 25.7h & 26.02 & 14M   & 24.7h \\
    s-MLP & 31.14 & \textbf{11M} & \textbf{1.4h} & \textbf{25.61} & \textbf{11M} & \textbf{2.02h} & \textbf{27.28} & \textbf{11M} & \textbf{1.87h} \\
    \bottomrule
    \end{tabular}%

%% file: tables/finetune_all_scene.tex
    \begin{tabular}{l|ccc|ccc|ccc|ccc|ccc|ccc}
    \toprule
          & \multicolumn{6}{c|}{Synthetic-NeRF}           & \multicolumn{6}{c|}{LLFF}                     & \multicolumn{6}{c}{TanksAndTemples} \\
    \midrule
          & \multicolumn{3}{c|}{t-Hash (Case 1)} & \multicolumn{3}{c|}{t-Tesnors (Case 2)} & \multicolumn{3}{c|}{t-Hash (Case 1)} & \multicolumn{3}{c|}{t-Tesnors (Case 2)} & \multicolumn{3}{c|}{t-Hash (Case 1)} & \multicolumn{3}{c}{t-Tesnors (Case 2)} \\
    \midrule
          & PSNR  & SSIM  & LPIPS$_{alex}$ & PSNR  & SSIM  & LPIPS$_{alex}$ & PSNR  & SSIM  & LPIPS$_{alex}$ & PSNR  & SSIM  & LPIPS$_{alex}$ & PSNR  & SSIM  & LPIPS$_{alex}$ & PSNR  & SSIM  & LPIPS$_{alex}$ \\
          & 32.58 & 0.96  & 0.032 & 27.49 & 0.917 & 0.122 & 26.7  & 0.832 & 0.13  & 21.33 & 0.59  & 0.53  & 29.26 & 0.915 & 0.106 & 25    & 0.865 & 0.279 \\
    \midrule
    s-VM  & \textbf{31.83} & \textbf{0.957} & 0.038 & 27.48 & 0.916 & 0.121 & 25.98 & 0.793 & \textbf{0.19} & 21.39 & 0.591 & 0.529 & \textbf{28.3} & 0.906 & \textbf{0.153} & 24.96 & 0.865 & 0.28 \\
    s-VM$_{ft}$ & 31.72 & 0.956 & \textbf{0.037} & \textbf{31.5} & \textbf{0.955} & \textbf{0.042} & \textbf{26.11} & \textbf{0.795} & 0.192 & \textbf{25.25} & \textbf{0.77} & \textbf{0.194} & 28.22 & 0.906 & 0.156 & \textbf{27.2} & \textbf{0.892} & \textbf{0.186} \\
    \midrule
    s-MLP & \textbf{31.63} & \textbf{0.953} & \textbf{0.043} & 27.38 & 0.914 & 0.119 & \textbf{25.95} & 0.786 & 0.201 & 21.33 & 0.592 & 0.53  & \textbf{27.97} & \textbf{0.9} & \textbf{0.152} & 24.78 & 0.863 & 0.273 \\
    s-MLP$_{ft}$ & 31.46 & 0.95  & 0.046 & \textbf{30.81} & \textbf{0.948} & \textbf{0.045} & 25.84 & \textbf{0.788} & 0.201 & \textbf{25.04} & \textbf{0.75} & \textbf{0.22} & 27.93 & 0.896 & 0.155 & \textbf{25.89} & \textbf{0.877} & \textbf{0.205} \\
    \bottomrule
    \end{tabular}%

%% file: tables/ablation.tex
    \begin{tabular}{l|cccc}
    \toprule
    \multirow{2}[2]{*}{} & \multirow{2}[2]{*}{PSNR} & \multirow{2}[2]{*}{SSIM} & \multirow{2}[2]{*}{$\text{Lpips}_{alex}$} & \multirow{2}[2]{*}{$\text{Lpips}_{vgg}$} \\
          &       &       &       &  \\
    \midrule
    w/o $\mathcal{L}_{2}^{v}$ & 25.87 & 0.858 & 0.208 & 0.202 \\
    w/o $\mathcal{L}_{2}^{\sigma}$ & 26.16 & 0.873 & 0.205 & 0.201 \\
    w/o $\mathcal{L}_{2}^{c}$ & 26.34 & 0.873 & 0.206 & 0.199 \\
    w/o $\mathcal{L}_{2}^{rgb}$ & 24.53 & 0.808 & 0.225 & 0.232 \\
    w/o sigma-restric & 25.19 & 0.849 & 0.214 & 0.207 \\
    w/o block-wise & 26.13 & 0.868 & 0.206 & 0.203 \\
    w/o poses$^{AL}$ & 26.34 & 0.876 & 0.204 & 0.200 \\
    w/o rays$^{AL}$ & 26.03 & 0.869 & 0.205 & 0.207 \\
    w/o points$^{AL}$ & 26.52 & 0.881 & \textbf{0.201} & 0.199 \\
    w/all & \textbf{26.71} & \textbf{0.887} & \textbf{0.201} & \textbf{0.197} \\
    \bottomrule
    \end{tabular}%

%% file: tables/division_mlp.tex
    \begin{tabular}{c|ccccc}
    \toprule
          & K=2   & K=3   & K=4   & K=5   & K=6 \\
    \midrule
    PSNR  & 31.35 & 31.55 & 31.63  & 31.69 & 31.70 \\
    Time  & 1.35h & 1.39h & 1.42h & 1.47h & 1.53h \\
    \bottomrule
    \end{tabular}%